\documentclass{article}
\usepackage{amsfonts, amssymb, amsmath, amsthm, mathtools}
\usepackage{dsfont}
\usepackage{geometry}
\usepackage{booktabs}
\usepackage[round]{natbib}
\usepackage[utf8]{inputenc}
\usepackage{enumitem}
\usepackage{endnotes}
\usepackage{bbm}
\usepackage{comment}
\usepackage{xtab}
\usepackage{booktabs}  
\usepackage{array}     
\usepackage[colorinlistoftodos]{todonotes}
\newcolumntype{C}{>{\centering\arraybackslash}p{1.8cm}}  

\usepackage{graphicx}
\usepackage{xcolor}
\usepackage{caption}
\usepackage{hyperref}
\usepackage{algorithm}
\usepackage{algpseudocode}
\usepackage{multirow}
\hypersetup{
	colorlinks=true,
	linktoc=all,
	linkcolor=blue,
	urlcolor=blue,
	citecolor=blue,
	filecolor=blue,
	linktocpage=true
}
\usepackage{cleveref}

\newtheorem{theorem}{Theorem}

\makeatother
\let\footnote=\endnote

\newcommand{\given}{\,|\,}

\title{Learning Majority-to-Minority Transformations with MMD and Triplet Loss for Imbalanced Classification}

\author{
Suman Cha \and
Hyunjoong Kim}

\date{\today}

\begin{document}
\maketitle

\begin{abstract}
Class imbalance in supervised classification often degrades model performance by biasing predictions toward the majority class, particularly in critical applications such as medical diagnosis and fraud detection. Traditional oversampling techniques—including SMOTE and its variants—generate synthetic minority samples via local interpolation but fail to capture global data distributions in high-dimensional spaces. Deep generative models based on GANs offer richer distribution modeling yet suffer from training instability and mode collapse under severe imbalance.
To overcome these limitations, we introduce an oversampling framework that learns a parametric transformation to map majority samples into the minority distribution. Our approach minimizes the maximum mean discrepancy (MMD) between transformed and true minority samples for global alignment, and incorporates a triplet loss regularizer to enforce boundary awareness by guiding synthesized samples toward challenging borderline regions. We evaluate our method on 29 synthetic and real-world datasets, demonstrating consistent improvements over classical and generative baselines in AUROC, G-mean, F1-score, and MCC. These results confirm the robustness, computational efficiency, and practical utility of the proposed framework for imbalanced classification tasks.
\end{abstract}

\section{Introduction}
Class imbalance is a pervasive challenge in supervised classification where minority class instances are substantially outnumbered by majority class samples \citep{he2009learning, krawczyk2016learning}. This problem manifests in critical domains including medical diagnosis \citep{roy2024learning}, fraud detection \citep{awoyemi2017credit}, and anomaly detection \citep{chandola2009anomaly}, where poor recognition of minority cases can yield severe consequences \citep{japkowicz2002class}. Aside from impairing predictive performance, class imbalance fosters algorithmic bias, leading to inaccurate representations and distorted decision-making \citep{binns2018fairness, mehrabi2021survey}. 

Data-level resampling, particularly oversampling the minority class, remains a prominent strategy for mitigating imbalance \citep{drummond2003c4}. Traditional methods such as the Synthetic Minority Oversampling Technique (SMOTE) \citep{chawla2002smote} and its variants including Borderline-SMOTE \citep{han2005borderline} and ADASYN \citep{he2008adasyn} synthesizes minority samples via local interpolations among existing minority instances. While simple and computationally efficient, these methods struggle to capture the global data distribution, especially in high-dimensional feature spaces \citep{bellinger2016beyond, alkhawaldeh2023challenges}, often producing samples that are noisy or situated far from decision boundaries. 

More recently, deep generative models have been explored to overcome these shortcomings. Methods based on Generative Adversarial Networks (GANs) \citep{goodfellow2014generative} and Variational Autoencoders (VAEs) \citep{kingma2013auto} offer richer minority samples generation reflecting complex data distributions. For instance, Generative Adversarial Minority Oversampling (GAMO) \citep{mullick2019generative} introduces an auxiliary classifier to encourage boundary-aware sample synthesis. BAGAN \citep{mariani2018bagan} further alleviates imbalance by jointly training on all classes and employing an autoencoder-based initialization to learn class-conditioned latent representations, enabling the generator to produce diverse and high-quality minority examples. SMOTified-GAN \citep{sharma2022smotified} incorporates SMOTE principles into GANs to address minority class sparsity, while VAE-based approaches \citep{zhang2018over} use the latent generative capacities to produce diverse minority samples. Despite these advances, deep generative model-based methods remain challenging due to training instability, mode collapse, and hyperparameter sensitivity, particularly under extreme imbalance settings \citep{salimans2016improved}. 

Another vein of research has focused on exploiting abundant majority data to guide synthetic minority generation. MWMOTE \citep{barua2012mwmote} prioritizes majority class regions for oversampling through local instance weighting. The M2m framework \citep{kim2020m2m} introduces a classifier-dependent transformation from majority samples to the minority domain, but relies on pretrained classifier to define decision boundary. Majority-Guided VAE (MGVAE) \citep{ai2023generative} transfers majority diversity via structured latent priors but confines guidance to latent space without explicit alignment to input-space decision boundaries. 

Motivated by these limitations, we propose a novel oversampling framework that learns an explicit parametric mapping $f_\theta$ to transform majority samples into the minority distribution. Our method minimizes the Maximum Mean Discrepancy (MMD) \citep{gretton2012kernel} between transformed and true minority samples to ensure global distributional alignment. To guarantee that generated samples reside in informative regions near the decision boundary, we introduce a triplet loss regularizer \citep{weinberger2009distance, schroff2015facenet}. We define triplets inspired by the danger set concept in Borderline-SMOTE \citep{han2005borderline}. This regularizer explicitly pushes transformed samples towards borderline minority points while maintaining separation from majority clusters.

The key contributions of this paper are twofold. First, we propose a novel oversampling framework that learns a parametric majority-to-minority transformation by minimizing the MMD. Second, we introduce a boundary-aware triplet regularizer that promotes sample generation near decision boundaries. Extensive experiments on both synthetic and real-world datasets demonstrate that our method consistently outperforms existing oversampling techniques across various imbalance settings.

The remainder of this paper is organized as follows. Section~\ref{sec:related_works} provides a comprehensive review of related literature. Section~\ref{sec:proposed_method} details the proposed approach. Section~\ref{sec:simulation_studies} and Section~\ref{real_data_exp} present empirical results from simulations and real data. Finally, Section~\ref{sec: conclusion} offers concluding remarks and outlines avenues for future research.

\section{Related Works} \label{sec:related_works}
Synthetic sample generation to address class imbalance has attracted significant attention in machine learning. This section reviews relevant prior work emphasizing three interconnected research paradigms: kernel-based distributional alignment via MMD, boundary-aware geometric regularization employing triplet loss, and integrated oversampling strategies that simultaneously address both global distributional properties and local boundary sensitivity. 

Integral probability metrics, particularly MMD, have become foundational tools across tasks including two-sample testing \citep{gretton2012kernel, liu2020learning, kubler2020learning}, domain adaptation \citep{long2015learning, tzeng2017adversarial}, and deep generative modeling \citep{dziugaite2015training, li2015generative}. Within generative modeling, MMD offers a kernel-based discrepancy measure that promotes alignment between generated and target distributions. Generative Moment Matching Networks (GMMN) \citep{li2015generative} pioneered this approach by replacing the discriminator in GANs with a two-sample test using MMD. This replacement facilitates stable generator training without adversarial objectives. To improve model expressiveness and training efficiency, subsequent work such as MMD-GAN \citep{li2017mmd} introduced kernel adversarial learning. Despite these advancements, existing MMD-based methods have been applied primarily to image synthesis and unsupervised generation tasks, with limited focus on class imbalance scenarios.
using the statistical consistency and stable convergence of MMD \citep{binkowski2018demystifying}, we adopt it as a principled distributional loss to circumvent adversarial training pitfalls—such as mode collapse and gradient instability \citep{salimans2016improved}.

Geometric regularization frameworks based on triplet loss \citep{weinberger2009distance, schroff2015facenet}, have demonstrated substantial success in metric learning \citep{hoffer2015deep, sohn2016improved} and discriminative representation learning for applications including face recognition \citep{schroff2015facenet} and image retrieval \citep{hermans2017defense}. In imbalanced classification, triplet loss has primarily been used to enhance feature-space separability at the classifier level \citep{kang2019decoupling, li2020overcoming}, with its potential to guide synthetic sample generation remaining largely unexplored. Traditional oversampling methods such as Borderline-SMOTE \citep{han2005borderline} and MWMOTE \citep{barua2012mwmote} incorporate heuristic boundary-awareness, but lack formal geometric regularization mechanisms that explicitly encode relative distances among synthetic, minority, and majority samples.

Only a small number of recent studies have  attempted to integrate both distributional alignment and geometric placement in oversampling. The M2m framework \citep{kim2020m2m} learns a classifier-guided mapping from majority to minority domains but omits explicit distributional discrepancy criteria and relies on the performance of the classifier. Majority-guided VAE (MGVAE) \citep{ai2023generative} incorporates majority class structure through latent priors, operating primarily in the latent representation space without explicit control of decision boundary alignment. Likewise, adversarial generative oversampling models including BAGAN \citep{mariani2018bagan} and SMOTified-GAN \citep{sharma2022smotified} focus on sample fidelity and global feature alignment but do not impose boundary-aware regularization. 

In summary, while MMD-based distribution matching and triplet-loss geometric regularization have been independently applied in various machine learning settings, their combined use for imbalanced classification remains unexplored. To our knowledge, this paper presents the first transport-map framework that integrates MMD-based distribution alignment with boundary-aware triplet regularization for synthetic minority sample generation. Beyond advancing oversampling methods for tabular data, our approach establishes a theoretical foundation adaptable to more complex modalities.

\section{Proposed Method}\label{sec:proposed_method}
In this section, we introduce a transport-map-based oversampling framework that learns a parametric transformation to project majority samples into the minority class domain. This framework combines global distribution alignment via MMD with local geometric regularization employing a triplet loss to ensure that generated samples populate informative boundary regions. We formalize the problem and elaborate on the two loss components below. 

\subsection{Problem Formulation}
Consider a labeled dataset $\mathcal{D} = \{(\mathbf{x}_i, y_i)\}_{i=1}^{n}$, where $\mathbf{x}_i \in \mathbb{R}^d$ and $y_i \in \{0,1\}$, where $0$ denotes the majority class and $1$ the minority class. We partition $\mathcal{D}$ into majority $\mathcal{D}_{\text{maj}} = \{\mathbf{x}_i \given y_i = 0\}$ of size $n_{\text{maj}}$ and minority $\mathcal{D}_{\text{min}} = \{\mathbf{x}_i \given y_i = 1\}$ of size $n_{\text{min}}$, where $n_{\text{maj}} \gg n_{\text{min}}$. Our objective is to learn a mapping $f_\theta : \mathbb{R}^d \to \mathbb{R}^d$ such that the distribution of transformed majority samples $f_\theta(\mathcal{D}_{\text{maj}})$ approximates the true minority distribution. \noindent Formally, we solve
\begin{equation*}
\theta^{\ast} = \arg\min_\theta \mathcal{L}_{\text{total}}(f_\theta(\mathcal{D}_\text{maj}), \mathcal{D}_\text{min}),     
\end{equation*}
with the total loss decomposed as  
\begin{equation} \label{eq: total_obj}
    \mathcal{L}_{\text{total}}(f_\theta(\mathcal{D}_{\text{maj}}), \mathcal{D}_{\text{min}}) = \mathcal{L}_{\text{MMD}}(f_\theta(\mathcal{D}_{\text{maj}}), \mathcal{D}_{\text{min}}) + \lambda \mathcal{L}_{\text{triplet}}(f_\theta(\mathcal{D}_{\text{maj}}), \mathcal{D}_{\text{min}}),
\end{equation}
where $\lambda > 0$ trades off global distributional alignment and boundary-aware regularization. After training the mapping, synthetic minority samples are generated by applying the learned transformation to majority instances as $\tilde{\mathbf{x}} = f_{\theta^{\ast}}(\mathbf{x})$ for $\mathbf{x} \in \mathcal{D}_{\text{maj}}$.  

\vspace{1em}
\noindent 
\textbf{Remark. } In our study, we implemented the transformation map $f_\theta : \mathbb{R}^d \rightarrow \mathbb{R}^d$ as a deep neural network following a residual connection-based autoencoder architecture, which empirically facilitated stable training and flexible modeling capacity. However, the proposed framework is not restricted to this architectural choice. The essential requirement is merely that $f_\theta$ be a measurable mapping from $\mathbb{R}^d$ to $\mathbb{R}^d$, as demanded by the distribution-matching criterion imposed by the MMD loss. Thus, in principle, $f_\theta$ may be instantiated as any sufficiently expressive parametric or nonparametric function—such as feedforward networks, invertible flows, kernel regressors, or even shallow linear maps—provided they admit optimization via the chosen loss in Equation~\eqref{eq: total_obj}. This generality enables practitioners to adapt $f_\theta$ to the structural properties or inductive biases most appropriate for a given application or data modality, without altering the fundamental learning objective.

\subsection{Maximum Mean Discrepancy} 
To ensure the distribution of transformed samples align with the minority distribution, we minimize the MMD. MMD is a kernel-based integral probability metric measuring the distance between two probability measures $P$ and $Q$ via mean embeddings in a Reproducing Kernel Hilbert Space (RKHS), $\mathcal{H}_k$. For a characteristic kernel $k(\cdot, \cdot)$, and associated feature map $\phi: \mathbb{R}^d \mapsto \mathcal{H}_k$, the squared MMD is
\begin{equation}\label{equation:MMD_norm}
\text{MMD}^2(P, Q) = \|\mathbb{E}_{\mathbf{x} \sim P}[\phi(\mathbf{x})] - \mathbb{E}_{\mathbf{z} \sim Q}[\phi(\mathbf{z})]\|_{\mathcal{H}}^2.    
\end{equation}
By applying the kernel trick, Equation~\eqref{equation:MMD_norm} can be expanded into a more tractable form without explicitly defining the feature map $\phi$. 
\begin{equation}
    \text{MMD}^2(P,Q) = \mathbb{E}_{\mathbf{x}, \mathbf{x}' \sim P}[k(\mathbf{x}, \mathbf{x}')] + \mathbb{E}_{\mathbf{z}, \mathbf{z}' \sim Q}[k(\mathbf{z}, \mathbf{z}')] - 2\mathbb{E}_{\mathbf{x} \sim P, \mathbf{z} \sim Q}[k(\mathbf{x}, \mathbf{z})], 
\end{equation}
where $\mathbf{x}'$ and $\mathbf{z}'$ are independent copies of $\mathbf{x}$ and $\mathbf{z}$, respectively. A key property of MMD, which forms the theoretical foundation of our approach is its ability to uniquely identify distributions when a characteristic kernel is used. This property is formally stated in the following theorem from \citet{gretton2012kernel}.

\vspace{0.8em}
\noindent
\begin{theorem} [\citet{gretton2012kernel}] Let $\mathcal{H}$ be a RKHS on a compact metric space $\mathcal{X}$ with a characteristic kernel $k(\cdot, \cdot)$. Then for any two Borel probability measures $P$ and $Q$ on $\mathcal{X}$, $\text{MMD}(P, Q) = 0$ if and only if $P=Q$.
\end{theorem}
\vspace{0.8em}

This theorem guarantees that by minimizing the MMD between the distribution of transformed majority samples and the true minority distribution, we are effectively driving the two distributions to become identical. Given the finite sample sets $\mathcal{D}_{\text{maj}}$ and $\mathcal{D}_{\text{min}}$, we employ the biased empirical estimator of the squared MMD as our loss function:
\begin{equation*}
    \mathcal{L}_{\text{MMD}} = \frac{1}{n_{\text{maj}}^2}\sum_{i,j=1}^{n_{\text{maj}}}k\left(f_{\theta}(\mathbf{x}_i), f_{\theta}(\mathbf{x}_{j})\right) + \frac{1}{n_{\text{min}}^2}\sum_{i,j=1}^{n_{\text{min}}}k(\mathbf{z}_i, \mathbf{z}_{j}) - \frac{2}{n_{\text{maj}}n_{\text{min}}}\sum_{i=1}^{n_{\text{maj}}}\sum_{j=1}^{n_{\text{min}}}k(f_\theta(\mathbf{x}_i), \mathbf{z}_j),
\end{equation*}
where $\mathbf{x}_i \in \mathcal{D}_{\text{maj}}$ and $\mathbf{z}_j \in\mathcal{D}_{\text{min}}$. In our experiments, we employ the Gaussian kernel $k(\mathbf{u}, \mathbf{v}) = \exp\left(-\|\mathbf{u} - \mathbf{v}\|_2^2/2\sigma^2 \right)$, with the bandwidth parameter $\sigma$ selected using the median heuristic.

\subsection{Triplet Loss Regularizer}
While MMD ensures global alignment of transformed majority samples to the minority distribution, it does not guarantee that generated samples populate regions near the decision boundary, which are crucial for effective classifier learning. To address this limitation, we incorporate a triplet loss \citep{schroff2015facenet} constructed over anchors, positives, and negatives. 

In our setting, anchors correspond to transformed majority samples $\tilde{\mathbf{x}} = f_\theta(\mathbf{x})$ where $\mathbf{x} \in \mathcal{D}_{\mathrm{maj}}$. Inspired by the methodology of Borderline-SMOTE \citep{han2005borderline}, we define \emph{danger set} as positives $\mathbf{x}^+$. The danger set is defined by minority samples having more than $k/2$ but fewer than $k$ majority neighbors among their $k$-nearest neighbors. Negatives $\mathbf{x}^-$ corresponds to \emph{safe set} containing majority instances deep inside the majority manifold, thus distant from the decision boundary. The triplet loss enforces that the anchor is closer to the positive than the negative by a margin $\alpha > 0$, formally, 
\begin{equation*}
    \mathcal{L}_{\text{triplet}} = \max\left(\|\tilde{\mathbf{x}} - \mathbf{x}^+\|_2^2 - \|\tilde{\mathbf{x}} - \mathbf{x}^-\|_2^2 + \alpha, 0\right).
\end{equation*}

Minimizing $\mathcal{L}_{\mathrm{triplet}}$ encourages $f_\theta$ to generate synthetic samples that concentrate near minority boundary points while maintaining separation from safe majority regions. This regularization increases the informativeness of generated samples, thereby enhancing classifier robustness and generalization.

\section{Simulation Studies}\label{sec:simulation_studies}
In this section, we provide a qualitative assessment of our proposed oversampling framework through visual analysis on two synthetic datasets. We consider two scenarios: (i) a Gaussian Blob dataset and (ii) a Half-moon dataset.

\begin{figure}[ht!]
\centering
\includegraphics[width=0.95\linewidth]{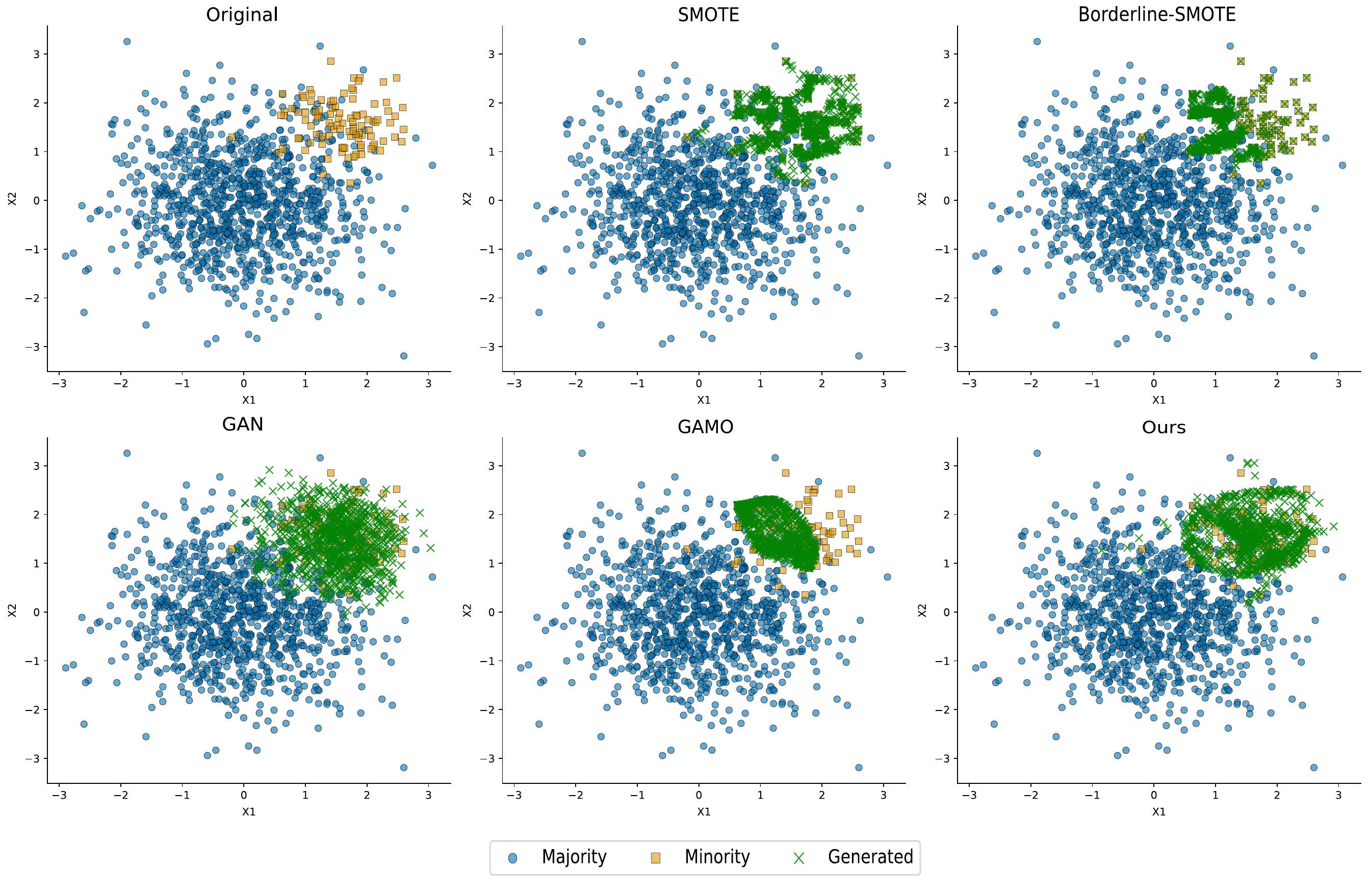}
\caption{Visualization of oversampling on the Gaussian Blob dataset. The majority (blue) and minority (orange) classes exhibit overlap. The samples generated by each method are represented in green. Our method generates samples that align well with the target minority distribution while effectively populating the decision boundary.}
\label{fig:gaussian_blob}
\end{figure}

\paragraph{Gaussian Blob.} This simulation evaluates a method's capacity to generate meaningful samples in a scenario with overlapping class distributions. The majority and minority classes, $D_{\text{maj}}$ and $D_{\text{min}}$, are drawn from two-dimensional Gaussian distributions. Specifically, the majority class follows $N(\mu_0, 0.5I_2)$ with $\mu_0 = (0, 0)$, while the minority class follows $N(\mu_1, I_2)$ with $\mu_1 = (0.5, 0.5)$, where $I_2$ is a two-dimensional identity matrix. This setup creates a region of class overlap, which presents a challenge for oversampling methods that must generate samples that are not only distributionally aligned but also located in informative, boundary-proximal regions. The objective is to evaluate whether the learned transformation can effectively generate minority-class samples while preserving the data’s intrinsic geometric structure. As shown in Figure~\ref{fig:gaussian_blob}, our method successfully transforms majority samples into the minority cluster, generating synthetic instances that are well-aligned with the target distribution. The MMD loss ensures that the generated samples cover the full spread of the minority class, while the triplet loss regularizer steers them toward the decision boundary.

\begin{figure}[ht!]
\centering
\includegraphics[width=0.95\linewidth]{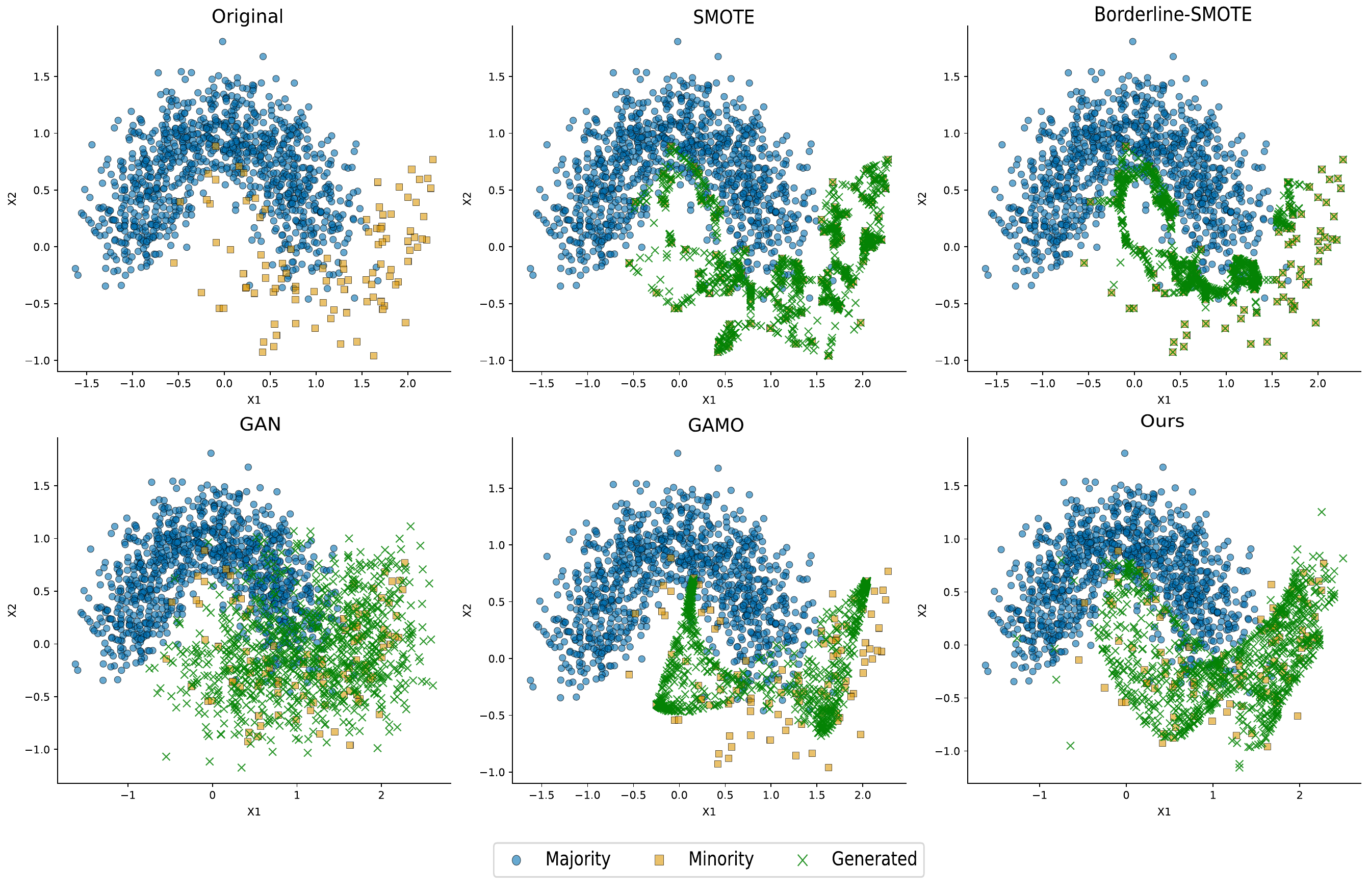}
\caption{Visualization of oversampling on the Half-moon dataset, a standard benchmark for evaluating performance on non-linear manifolds. The samples generated by each method are represented in green. Our method successfully learns the required non-linear transformation, generating synthetic samples that conform to the geometric structure of the minority class.}
\label{fig:half_moon}
\end{figure}

\paragraph{Half-moon.} The Half-moon dataset is employed to assess the framework's ability to learn a highly non-linear transformation. The data consists of two interleaving half-moon shapes, with a noise level of 0.25 introduced to create ambiguity along the class boundary. This configuration tests the model's flexibility in mapping a geometrically simple distribution onto a complex, non-convex manifold. As depicted in Figure~\ref{fig:half_moon}, our method demonstrates a strong capability to learn the required non-linear mapping. The generated samples conform closely to the crescent shape of the minority class, indicating that the transformation successfully captures the underlying data manifold. The integration of MMD ensures that the density of the generated samples matches that of the original minority moon, while the triplet loss places these samples in regions most beneficial for defining the non-linear decision boundary. Other generative models often fail to capture complex distributions under data sparsity, yielding incoherent samples; this highlights the advantage of our transport-map–based approach, which uses abundant majority-class information to guide the generation process.

\section{Real data Experiments}
\label{real_data_exp}
In this section, we evaluate our proposed oversampling framework on real-world imbalanced datasets to demonstrate its effectiveness in enhancing classification performance under varying degrees of class imbalance. We assess the effectiveness of our method in comparison to established baselines across various performance metrics. 

\subsection{Experimental Setup}\label{subsec:setup}
We evaluate the proposed method on 29 real-world imbalanced datasets commonly used in the literature \citep{chawla2002smote, barua2012mwmote}. Table~\ref{tab:dataset_summary} summarizes key characteristics of these datasets from diverse domains, including total sample size ($N_{\text{total}}$), minority sample size ($N_{\text{min}}$), imbalance ratio (IR), and feature dimension ($d$). Their wide range of IR values provides a robust testbed for evaluating oversampling methods across varying degrees of class imbalance.

\begin{table*}[ht!]
\centering
\caption{Summary of 29 real-world imbalanced datasets employed in experiments. Each dataset is assigned a shorthand identifier ($D_1$--$D_{29}$) for clarity in subsequent result tables. Data characteristics include total size ($N_{\mathrm{total}}$), minority size ($N_{\mathrm{min}}$), imbalance ratio (IR), and feature dimension ($d$).}
\label{tab:dataset_summary}
\begin{tabular}{c c c c c c}
\toprule
ID & Dataset Name & $N_{\text{total}}$ & $N_{\text{min}}$ & IR & $d$ \\
\midrule
$D_{1}$ & \texttt{abalone19} & 4174 & 31 & 129.44 & 8 \\
$D_{2}$ & \texttt{abalone9-18} & 731 & 42 & 16.4 & 8 \\
$D_{3}$ & \texttt{arrhythmia} & 452 & 25 & 17 & 278 \\
$D_{4}$ & \texttt{cleveland-0\_vs\_4} & 177 & 12 & 12.62 & 13 \\
$D_{5}$ & \texttt{coil\_2000} & 9822 & 577 & 16 & 85 \\
$D_{6}$ & \texttt{ecoli3} & 336 & 35 & 8.6 & 7 \\
$D_{7}$ & \texttt{ecoli4} & 336 & 20 & 15.8 & 7 \\
$D_{8}$ & \texttt{glass-0-1-2-3\_vs\_4-5-6} & 214 & 50 & 3.2 & 9 \\
$D_{9}$ & \texttt{glass-0-1-6\_vs\_5} & 184 & 9 & 19.44 & 9 \\
$D_{10}$ & \texttt{glass4} & 214 & 12 & 15.47 & 9 \\
$D_{11}$ & \texttt{oil} & 937 & 40 & 22 & 49 \\
$D_{12}$ & \texttt{ozone\_level} & 2536 & 72 & 34 & 72 \\
$D_{13}$ & \texttt{pima} & 768 & 267 & 1.87 & 8 \\
$D_{14}$ & \texttt{scene} & 2407 & 171 & 13 & 294 \\
$D_{15}$ & \texttt{segment0} & 2308 & 328 & 62 & 19 \\
$D_{16}$ & \texttt{sick\_euthyroid} & 3163 & 292 & 9.8 & 42 \\
$D_{17}$ & \texttt{solar\_flare\_m0} & 1389 & 69 & 19 & 32 \\
$D_{18}$ & \texttt{thyroid\_sick} & 3772 & 235 & 15 & 52 \\
$D_{19}$ & \texttt{us\_crime} & 1994 & 153 & 12 & 100 \\
$D_{20}$ & \texttt{wine\_quality} & 4898 & 181 & 26 & 11 \\
$D_{21}$ & \texttt{winequality-red-3\_vs\_5} & 691 & 10 & 68.1 & 11 \\
$D_{22}$ & \texttt{winequality-red-4} & 1599 & 52 & 29.17 & 11 \\
$D_{23}$ & \texttt{winequality-white-3\_vs\_7} & 900 & 20 & 44 & 11 \\
$D_{24}$ & \texttt{winequality-white-3-9\_vs\_5} & 1482 & 25 & 58.28 & 11 \\
$D_{25}$ & \texttt{yeast-0-5-6-7-9\_vs\_4} & 528 & 51 & 9.35 & 8 \\
$D_{26}$ & \texttt{yeast-1-2-8-9\_vs\_7} & 947 & 29 & 30.57 & 8 \\
$D_{27}$ & \texttt{yeast-1-4-5-8\_vs\_7} & 693 & 30 & 22.1 & 8 \\
$D_{28}$ & \texttt{yeast4} & 1484 & 50 & 28.1 & 8 \\
$D_{29}$ & \texttt{yeast6} & 1484 & 35 & 41.4 & 8 \\
\bottomrule
\end{tabular}
\end{table*}

We benchmark our method against a comprehensive set of state-of-the-art oversampling algorithms, each representing a distinct approach to imbalanced classification. For each baseline, minority class oversampling is applied solely to the training data until class balance is reached, prior to classifier training. Unless otherwise specified, a Support Vector Machine (SVM) with a radial basis function (RBF) kernel serves as the base learner, consistent with established protocols in imbalanced learning research \citep{krawczyk2016learning}. Default hyperparameters are adopted for all methods, following recommendations in their respective original works. The compared methods include: 

\begin{itemize}[leftmargin=1.25em]
    \item \textbf{Random OverSampling (ROS)} \citep{he2009learning}: Uniformly duplicates minority instances until class balance is achieved., introducing no synthetic variation but serving as a canonical baseline for assessing the impact of sample size alone. 
    \item \textbf{SMOTE} \citep{chawla2002smote}: Produces synthetic minority examples by linearly interpolating between a minority instance and its k-nearest minority neighbors, thereby increasing within-class diversity and expanding the minority support. 
    \item \textbf{Borderline-SMOTE (bSMOTE)} \citep{han2005borderline}: Focuses synthetic generation on so-called danger samples near the decision boundary, as determined by local neighborhood structure, to augment boundary informativeness. 
    \item \textbf{ADASYN} \citep{he2008adasyn}: Allocates more synthetic generation to minority samples surrounded by a larger fraction of majority class neighbors, adaptively addressing local imbalance in the feature space.
    \item \textbf{MWMOTE} \citep{barua2012mwmote}: Employs an instance-weighting scheme to prioritize minority samples in regions deemed most informative for classification, based on proximity to the majority class and cluster structure. 
    \item \textbf{CTGAN} \citep{xu2019modeling}: uses a conditional generative adversarial network specifically designed for structured tabular data, enabling high-fidelity synthetic minority sample generation by modeling conditional distribution. 
    \item \textbf{GAMO} \citep{mullick2019generative}: Augments adversarial learning with an auxiliary classifier to encourage generation of synthetic minority samples that are decision-boundary-aware. 
    \item \textbf{MGVAE} \citep{ai2023generative}: Leverages a majority-guided variational autoencoder to transfer diversity from the majority to the minority class, synthesizing realistic minority data through a learned prior informed by majority characteristics. 
\end{itemize}

\vspace{1em}

\noindent Performance is evaluated using four widely adopted metrics that are considered suitable for class imbalanced learning, each characterizing different facets of performance with sensitivity to minority class behavior. The metrics are defined and interpreted as follows: 
\begin{itemize}[leftmargin=1.25em]
    \item \textbf{Area Under Receiver Operating Characteristic Curve (AUROC)} \citep{fawcett2006introduction}: Measures the probability that a randomly selected positive instance ranks higher than a randomly selected negative instance according to the classifier's predicted scores. Formally, for a scoring function $s(\cdot)$, AUROC computes
    \begin{equation*}
        \text{AUROC} = \mathbb{P}\left( s(x^+) > s(x^-)\right),
    \end{equation*}
    where $x^+ \sim \mathcal{D}_{\text{min}}$ and $x^- \sim \mathcal{D}_{\text{maj}}$. The metric is threshold-independent and robust to imbalance because it evaluates ranking rather than absolute predictions. 

\item \textbf{Geometric Mean (G-mean)} \citep{he2009learning}:  
Defined as the square root of the product of the true positive rate (TPR) and true negative rate (TNR), this metric penalizes classifiers that favor one class over the other. For binary classification,  
\[
   \mathrm{G\text{-}mean} 
   = \sqrt{\mathrm{TPR} \times \mathrm{TNR}}
   = \sqrt{\frac{\mathrm{TP}}{\mathrm{TP} + \mathrm{FN}} \times \frac{\mathrm{TN}}{\mathrm{TN} + \mathrm{FP}}}\,,
\]
where 
\begin{itemize}[nosep,leftmargin=1.25em]
  \item TP: true positives (correctly predicted positive instances),  
  \item FN: false negatives (positive instances incorrectly predicted as negative),  
  \item TN: true negatives (correctly predicted negative instances),  
  \item FP: false positives (negative instances incorrectly predicted as positive).  
\end{itemize}
A high G-mean indicates strong, balanced performance on both classes.

    \item \textbf{F1-score}: Captures the harmonic mean of precision and recall with the formula:
    \begin{equation*}
        \text{F1} = \frac{2\times \text{Precision} \times \text{Recall}}{\text{Precision} + \text{Recall}} = \frac{2\times\text{TP}}{2 \times \text{TP} + \text{FP} + \text{FN}}.
    \end{equation*}
It measures the balance between precision and recall for the minority class, placing equal emphasis on both false negatives and false positives. 

    \item \textbf{Matthews Correlation Coefficient (MCC)} \citep{chicco2020advantages}: Provides a correlation coefficient between the predicted and true labels, defined as:
    \begin{equation*}
        \text{MCC} = \frac{\text{TP} \times \text{TN} - \text{FP} \times \text{FN}}{\sqrt{(\text{TP}+ \text{FP}) \times (\text{TP} + \text{FN}) \times (\text{TN} + \text{FP}) \times (\text{TN} + \text{FP})}}.
    \end{equation*}
    $\text{MCC}$ ranges from $-1$ to $1$, where $1$ indicates perfect prediction, $0$ corresponds to random guessing, and $-1$ reflects complete disagreement.
\end{itemize}

\vspace{1em}
\noindent Evaluation is conducted through 10-fold cross-validation repeated over 10 independent trials, following the work of \citet{mostafaei2023ouboost}. All results are reported as mean values across folds, along with per-dataset ranks to facilitate comparative analysis. 

\subsection{Main Results}\label{subsec: main_results}

We present comparative results across 29 real-world benchmark datasets, evaluating the effectiveness of our proposed oversampling framework against both traditional and generative baselines. Performance is measured in terms of AUROC, G-mean, F1-score, and MCC as defined in Section~\ref{subsec:setup}. To ensure robust conclusions, we report average scores with their corresponding ranks and assess statistical significance via paired t-tests comparing our proposed method both to the Original (non-oversampled) baseline and to the best-performing alternative method.

\vspace{0.5em}
\noindent \textbf{Main Performance.} Tables~\ref{tab:auroc_results}–\ref{tab:mcc_results} summarize the average performance metrics of all methods, with ranks indicating their relative ordering for each dataset. Our method achieves the highest average rank among all baselines, demonstrating consistent superiority across datasets with diverse properties, including extreme imbalance ratios and high-dimensional feature spaces.

\begin{table}[ht!]
\centering
\caption{AUROC results (with ranks in parentheses) for each dataset and oversampling method across 29 benchmark datasets. Better performance is indicated by higher values and lower average ranks. The best-performing method for each dataset is shown in bold. ‘Rank 1(2)’ indicates the number of first- and second-place finishes, with the highest counts highlighted in bold.}
\label{tab:auroc_results}
\resizebox{\textwidth}{!}{
\begin{tabular}{c c c c c c c c c c c}
\toprule
 Dataset &          Original &          ROS &             SMOTE &            bSMOTE &            ADASYN &            MWMOTE &             CTGAN &              GAMO &             MGVAE &               Ours \\
\midrule
 $D_{1}$ &        0.75 (6) & 0.7874 (4) &      0.7897 (2) &      0.7843 (5) &       0.789 (3) &      0.7382 (8) &      0.5648 (9) &     0.5608 (10) &      0.7439 (7) &       \textbf{0.8078 (1)} \\
 $D_{2}$ &      0.8396 (9) & \textbf{0.9044 (1)} &      0.8942 (3) &      0.8766 (6) &       0.883 (4) &      0.8784 (5) &      0.8442 (8) &      0.838 (10) &      0.8531 (7) &       0.9004 (2) \\
 $D_{3}$ &      0.8804 (7) & \textbf{0.9189 (1)} &      0.9108 (3) &       0.898 (6) &      0.9121 (2) &      0.9039 (4) &      0.7939 (9) &     0.7006 (10) &      0.8099 (8) &       0.8996 (5) \\
 $D_{4}$ &      \textbf{0.9862 (1)} & 0.9834 (5) &      0.9844 (3) &      0.9831 (6) &      0.9831 (6) &      0.9828 (8) &      0.9769 (9) &     0.8972 (10) &       0.984 (4) &       0.9859 (2) \\
 $D_{5}$ &      0.6482 (7) & \textbf{0.7031 (1)} &      0.6805 (4) &      0.6805 (4) &      0.6809 (3) &      0.6676 (6) &      0.6176 (9) &     0.5971 (10) &      0.6447 (8) &       0.6998 (2) \\
 $D_{6}$ &      0.9532 (4) & 0.9542 (2) &       \textbf{0.956 (1)} &      0.9483 (9) &      0.9512 (5) &      0.9498 (7) &      0.9486 (8) &     0.9019 (10) &      0.9508 (6) &       0.9536 (3) \\
 $D_{7}$ &      0.9614 (7) & 0.9633 (4) &       0.968 (3) &     0.9237 (10) &      0.9585 (9) &      0.9594 (8) &       0.972 (2) &      0.9618 (5) &      0.9615 (6) &       \textbf{0.9765 (1)} \\
 $D_{8}$ &      0.9784 (8) & 0.9806 (3) &      0.9789 (5) &      0.9789 (6) &      0.9772 (9) &     0.9748 (10) &      0.9788 (7) &      \textbf{0.9846 (1)} &      0.9799 (4) &       0.9834 (2) \\
 $D_{9}$ &     0.9688 (10) & 0.9848 (5) &      0.9854 (4) &      0.9822 (6) &       0.986 (3) &      0.9766 (9) &      0.9892 (1) &      0.9804 (7) &      0.9777 (8) &       0.9868 (2) \\
$D_{10}$ &      0.9893 (4) & 0.9861 (8) &      0.9905 (2) &       0.989 (7) &      0.9898 (3) &      0.9893 (4) &     0.9612 (10) &      0.9809 (9) &      \textbf{0.9922 (1)} &        0.989 (6) \\
$D_{11}$ &      0.9169 (4) & 0.9118 (8) &      0.9132 (6) &      0.9309 (2) &      0.9127 (7) &      0.9144 (5) &      0.8963 (9) &     0.8883 (10) &       \textbf{0.932 (1)} &        0.922 (3) \\
$D_{12}$ &      0.8331 (9) & 0.8663 (6) &       0.869 (5) &      0.8927 (2) &      0.8659 (7) &      0.8717 (4) &      0.8466 (8) &     0.7211 (10) &       0.872 (3) &       \textbf{0.9059 (1)} \\
$D_{13}$ &      0.8291 (2) & 0.8281 (3) &      0.8253 (5) &      0.8181 (9) &      0.8215 (7) &      0.8181 (8) &      0.8237 (6) &     0.8124 (10) &      0.8267 (4) &       \textbf{0.8314 (1)} \\
$D_{14}$ &       0.768 (8) & 0.7824 (3) &      0.7791 (4) &      0.7719 (7) &       0.775 (5) &      0.7745 (6) &      0.7395 (9) &     0.6813 (10) &      0.7853 (2) &       \textbf{0.7895 (1)} \\
$D_{15}$ &      0.9996 (8) & 0.9999 (2) &      0.9999 (3) &      0.9997 (7) &      0.9997 (5) &      \textbf{0.9999 (1)} &      0.9996 (9) &     0.9954 (10) &      0.9997 (6) &       0.9998 (4) \\
$D_{16}$ &      0.9418 (7) & \textbf{0.9505 (1)} &      0.9495 (3) &      0.9445 (6) &      0.9498 (2) &      0.9448 (5) &      0.9311 (9) &     0.9116 (10) &      0.9395 (8) &       0.9455 (4) \\
$D_{17}$ &      0.6587 (7) & 0.6921 (2) &      0.6614 (6) &      0.6717 (4) &      0.6566 (8) &       0.648 (9) &      \textbf{0.6932 (1)} &     0.6163 (10) &       0.666 (5) &       0.6844 (3) \\
$D_{18}$ &      0.9412 (7) & 0.9546 (3) &      0.9552 (2) &      0.9466 (6) &      \textbf{0.9556 (1)} &      0.9475 (5) &      0.9322 (9) &     0.9222 (10) &      0.9399 (8) &        0.948 (4) \\
$D_{19}$ &      0.8655 (9) &  0.908 (3) &      0.9006 (4) &      0.9142 (2) &      0.8939 (6) &      0.9002 (5) &      0.8921 (7) &     0.8623 (10) &      0.8679 (8) &       \textbf{0.9182 (1)} \\
$D_{20}$ &      0.7706 (8) & \textbf{0.8209 (1)} &      0.8043 (4) &       0.805 (3) &      0.8024 (5) &      0.7806 (6) &      0.7745 (7) &     0.6929 (10) &      0.7519 (9) &       0.8154 (2) \\
$D_{21}$ &      0.6742 (7) & 0.7352 (4) &      0.6348 (9) &      0.6733 (8) &     0.6309 (10) &      0.6835 (6) &      \textbf{0.8169 (1)} &      0.7519 (2) &      0.6988 (5) &       0.7507 (3) \\
$D_{22}$ &      0.6759 (8) & 0.7444 (2) &      0.7194 (5) &      0.7302 (3) &      0.7203 (4) &      0.7192 (6) &      0.6908 (7) &      0.6614 (9) &     0.5764 (10) &       \textbf{0.7466 (1)} \\
$D_{23}$ &      \textbf{0.8129 (1)} & 0.6507 (7) &      0.6318 (9) &      0.7362 (5) &      0.6346 (8) &     0.6247 (10) &      0.7941 (2) &      0.7765 (3) &      0.7624 (4) &       0.7074 (6) \\
$D_{24}$ &      0.8804 (3) & 0.8489 (7) &      0.8681 (6) &      0.8923 (2) &      0.8782 (4) &     0.8423 (10) &      0.8473 (9) &      \textbf{0.9015 (1)} &      0.8484 (8) &       0.8691 (5) \\
$D_{25}$ &      0.8849 (3) & 0.8846 (4) &      0.8835 (5) &       0.882 (6) &      0.8769 (7) &      0.8716 (8) &      0.8467 (9) &      0.811 (10) &      \textbf{0.8884 (1)} &       0.8856 (2) \\
$D_{26}$ &      0.6609 (8) & 0.7318 (4) &      0.7109 (5) &      \textbf{0.7511 (1)} &      0.7071 (6) &      0.7023 (7) &      0.7369 (3) &      0.6249 (9) &     0.6134 (10) &       0.7466 (2) \\
$D_{27}$ &      0.6388 (7) & 0.6781 (3) &      0.6777 (4) &      \textbf{0.7018 (1)} &      0.6631 (6) &      0.6763 (5) &      0.6384 (8) &     0.5369 (10) &      0.5845 (9) &       0.6831 (2) \\
$D_{28}$ &      0.8419 (9) & 0.8959 (3) &      0.8958 (4) &      0.8976 (2) &       0.892 (5) &      0.8914 (6) &       0.848 (8) &     0.7337 (10) &      0.8539 (7) &       \textbf{0.9083 (1)} \\
$D_{29}$ &      0.8377 (9) & 0.9255 (6) &      0.9341 (3) &      \textbf{0.9374 (1)} &      0.9237 (7) &      0.9342 (2) &      0.9286 (5) &      0.768 (10) &      0.9058 (8) &       0.9333 (4) \\
\midrule
Rank 1(2) &             2 (1) &        5 (4) &             1 (3) &             3 (5) &             1 (2) &             2 (1) &             3 (2) &             3 (1) &             4 (1) &              \textbf{8 (9)} \\
Avg. Rank & 6.33 &          3.7 & 4.23 & 4.97 & 5.33 & 6.13 & 6.77 & 8.23 & 5.87 & \textbf{2.87} \\
\bottomrule
\end{tabular}
}
\end{table}

\begin{table}[ht!]
\centering
\caption{G-mean results (with ranks in parentheses) for each dataset and oversampling method across 29 benchmark datasets. Better performance is indicated by higher values and lower average ranks. The best-performing method for each dataset is shown in bold. ‘Rank 1(2)’ indicates the number of first- and second-place finishes, with the highest counts highlighted in bold.}
\label{tab:gmean_results}
\resizebox{\textwidth}{!}{
\begin{tabular}{c c c c c c c c c c c}
\toprule
 Dataset & Original & ROS & SMOTE & bSMOTE & ADASYN & MWMOTE & CTGAN & GAMO & MGVAE & Ours \\
\midrule
 $D_{1}$ &                 0.0 (7) &            0.62 (2) &        0.6157 (4) &        0.3072 (5) &          0.6199 (3) &        0.2957 (6) &             0.0 (7) &             0.0 (7) &           0.0 (7) & \textbf{0.6834 (1)} \\
 $D_{2}$ &                0.1684 (10) & \textbf{0.8058 (1)} &        0.7575 (3) &        0.7042 (6) &          0.7438 (4) &        0.7255 (5) &          0.3117 (9) &          0.4622 (8) &        0.5218 (7) &           0.786 (2) \\
 $D_{3}$ &                  0.0 (8) &          0.4859 (2) &        0.4669 (3) &        0.4348 (5) &          0.4599 (4) &        0.4139 (6) &             0.0 (8) &          0.1223 (7) &           0.0 (8) & \textbf{0.7659 (1)} \\
 $D_{4}$ &           0.2622 (9) &          0.7534 (3) &         0.667 (7) &        0.6247 (8) &           0.667 (6) &          0.73 (4) &          0.7681 (2) &          0.6847 (5) &       0.2596 (10) & \textbf{0.8111 (1)} \\
 $D_{5}$ &                  0.0013 (10) &  \textbf{0.592 (1)} &        0.5516 (3) &        0.5325 (5) &          0.5544 (2) &        0.4847 (6) &          0.0298 (8) &          0.1541 (7) &        0.0026 (9) &          0.5409 (4) \\
 $D_{6}$ &                     0.7289 (10) &          0.8805 (3) &        0.8717 (4) &        0.8687 (6) &          0.8663 (7) &        0.8689 (5) &          0.8821 (2) &           0.796 (8) &         0.766 (9) & \textbf{0.8844 (1)} \\
 $D_{7}$ &                     0.8273 (10) &          0.8408 (8) &        0.8497 (7) &        0.8302 (9) & \textbf{0.8709 (1)} &         0.864 (4) &          0.8699 (3) &          0.8703 (2) &         0.854 (6) &          0.8605 (5) \\
 $D_{8}$ &      0.9131 (10) &          0.9342 (6) &        0.9386 (4) &        0.9331 (7) &          0.9408 (3) &        0.9384 (5) &          0.9161 (8) & \textbf{0.9548 (1)} &        0.9157 (9) &          0.9501 (2) \\
 $D_{9}$ &            0.2697 (10) &          0.6233 (4) &        0.6051 (6) &        0.6051 (6) &          0.6145 (5) &        0.5951 (8) &          0.7276 (2) & \textbf{0.7914 (1)} &         0.528 (9) &          0.6536 (3) \\
$D_{10}$ &                      0.6669 (9) &          0.8932 (3) &        0.8766 (5) &        0.8802 (4) &          0.9106 (2) &          0.87 (7) &         0.6567 (10) &          0.8755 (6) &        0.8153 (8) & \textbf{0.9148 (1)} \\
$D_{11}$ &                         0.5166 (8) &          0.7528 (2) &        0.7194 (4) &        0.7124 (6) &          0.7321 (3) &        0.7144 (5) &          0.5022 (9) &          0.6303 (7) &       0.4841 (10) & \textbf{0.7663 (1)} \\
$D_{12}$ &                    0.0 (9) &           0.737 (2) &        0.7221 (3) &        0.6796 (6) &          0.7153 (4) &        0.6946 (5) &             0.0 (9) &          0.0833 (8) &        0.1676 (7) & \textbf{0.8116 (1)} \\
$D_{13}$ &                       0.6926 (10) &          0.7363 (6) &        0.7368 (5) &        0.7438 (2) &          0.7436 (3) &        0.7377 (4) &          0.7136 (8) &          0.7312 (7) &        0.6995 (9) & \textbf{0.7443 (1)} \\
$D_{14}$ &                       0.0477 (9) &          0.4673 (2) &        0.4487 (5) &        0.4408 (6) &            0.45 (4) &        0.4533 (3) &          0.043 (10) &          0.2066 (7) &        0.1675 (8) &  \textbf{0.693 (1)} \\
$D_{15}$ &                     0.9888 (8) &          0.9916 (3) &         0.991 (6) &        0.9917 (2) & \textbf{0.9924 (1)} &        0.9914 (4) &           0.989 (7) &         0.9826 (10) &         0.984 (9) &          0.9913 (5) \\
$D_{16}$ &             0.3563 (9) & \textbf{0.8981 (1)} &        0.8971 (2) &        0.8804 (6) &           0.895 (3) &         0.881 (5) &          0.7786 (7) &         0.3365 (10) &         0.537 (8) &          0.8913 (4) \\
$D_{17}$ &                  0.0 (9) & \textbf{0.5917 (1)} &        0.5338 (4) &        0.5327 (5) &          0.5381 (3) &        0.4273 (6) &          0.2657 (7) &          0.1423 (8) &           0.0 (9) &          0.5615 (2) \\
$D_{18}$ &                0.2186 (9) &          0.8876 (3) &        0.8906 (2) &        0.8713 (4) & \textbf{0.8943 (1)} &        0.8698 (5) &          0.6317 (7) &         0.1338 (10) &        0.5007 (8) &          0.8587 (6) \\
$D_{19}$ &                    0.4955 (10) &          0.8211 (2) &        0.7763 (3) &        0.7736 (4) &          0.7695 (5) &        0.7684 (6) &          0.5315 (8) &          0.5686 (7) &        0.5174 (9) & \textbf{0.8442 (1)} \\
$D_{20}$ &                0.0816 (10) & \textbf{0.7204 (1)} &        0.7123 (2) &        0.6792 (5) &          0.7117 (3) &        0.6737 (6) &          0.3764 (8) &           0.478 (7) &        0.1482 (9) &           0.684 (4) \\
$D_{21}$ &        0.0 (10) &          0.1975 (2) &         0.194 (3) &        0.1367 (8) &          0.1546 (6) &        0.1782 (5) &          0.1491 (7) & \textbf{0.4232 (1)} &          0.01 (9) &           0.186 (4) \\
$D_{22}$ &             0.0 (10) &          0.6172 (3) &         0.615 (4) &        0.5311 (6) &          0.6189 (2) &        0.5706 (5) &          0.1145 (8) &          0.1787 (7) &        0.0041 (9) & \textbf{0.6337 (1)} \\
$D_{23}$ &      0.0 (9) &          0.3578 (4) &        0.2759 (6) &        0.3747 (3) &          0.2486 (7) &        0.2165 (8) &          0.3246 (5) &          0.4041 (2) &           0.0 (9) &  \textbf{0.421 (1)} \\
$D_{24}$ &     0.0 (10) &          0.4229 (4) &        0.2827 (6) &        0.3942 (5) &          0.2679 (7) &        0.1189 (8) &          0.5748 (2) & \textbf{0.7899 (1)} &        0.0807 (9) &          0.4732 (3) \\
$D_{25}$ &        0.3565 (10) &          0.7622 (2) &        0.7434 (4) &        0.7171 (7) & \textbf{0.7648 (1)} &        0.7193 (6) &          0.6995 (8) &          0.7224 (5) &        0.5826 (9) &          0.7614 (3) \\
$D_{26}$ & 0.0 (10) &          0.6142 (2) &        0.5631 (5) &          0.54 (6) &          0.5769 (3) &        0.5637 (4) &          0.5291 (7) &          0.4868 (8) &        0.0455 (9) & \textbf{0.6531 (1)} \\
$D_{27}$ & 0.0 (10) &           0.494 (5) &        0.5295 (2) &        0.5023 (4) &          0.5231 (3) &         0.466 (6) &          0.2965 (7) &          0.1736 (8) &        0.1077 (9) & \textbf{0.5394 (1)} \\
$D_{28}$ & 0.0045 (10) &          0.7694 (2) &        0.7602 (3) &        0.7283 (6) &          0.7581 (4) &        0.7349 (5) &          0.6516 (7) &          0.5119 (8) &        0.4652 (9) & \textbf{0.7921 (1)} \\
$D_{29}$ & 0.3704 (10) & 0.8159 (4) & 0.8116 (5) & 0.7696 (7) & 0.7962 (6) & 0.8164 (3) & \textbf{0.8497 (1)} &          0.6347 (9) & 0.7266 (8) & 0.8459 (2) \\
\midrule
Rank 1 (2) & 0 (0) & 5 (10) & 0 (4) & 0 (2) & 4 (3) & 0 (1) & 1 (4) & 4 (3) & 0 (0) & \textbf{15 (4)} \\
Avg Rank & 9.1 &  3.17 & 4.13 & 5.47 &  3.77 & 5.23 &   6.53 &   6.13 & 8.27 &  \textbf{2.27} \\
\bottomrule
\end{tabular}
}
\end{table}

\begin{table}[ht!]
\centering
\caption{F1-score results (with ranks in parentheses) for each dataset and oversampling method across 29 benchmark datasets. Better performance is indicated by higher values and lower average ranks. The best-performing method for each dataset is shown in bold. ‘Rank 1(2)’ indicates the number of first- and second-place finishes, with the highest counts highlighted in bold.}
\label{tab:f1_results}
\resizebox{\textwidth}{!}{
\begin{tabular}{c c c c c c c c c c c}
\toprule
 Dataset &    Original &                 ROS &               SMOTE &              bSMOTE &              ADASYN &              MWMOTE &               CTGAN &                GAMO &               MGVAE &                Ours \\
\midrule
 $D_{1}$ &     0.0 (7) &          0.0469 (5) &          0.0547 (2) &          0.0497 (4) & \textbf{0.0551 (1)} &          0.0418 (6) &             0.0 (7) &             0.0 (7) &             0.0 (7) &          0.0516 (3) \\
 $D_{2}$ &  0.134 (10) & \textbf{0.4267 (1)} &          0.4089 (3) &          0.4216 (2) &          0.3758 (7) &          0.3991 (5) &           0.247 (9) &          0.3319 (8) &          0.3836 (6) &          0.4036 (4) \\
 $D_{3}$ &     0.0 (8) &          0.3964 (2) &          0.3821 (3) &          0.3527 (5) &          0.3764 (4) &           0.336 (6) &             0.0 (8) &          0.0375 (7) &             0.0 (8) & \textbf{0.4538 (1)} \\
 $D_{4}$ &  0.2467 (9) &          0.6987 (2) &           0.627 (5) &          0.5883 (7) &           0.627 (6) &            0.68 (3) &          0.6541 (4) &          0.3773 (8) &         0.2467 (10) & \textbf{0.7397 (1)} \\
 $D_{5}$ & 0.0003 (10) &          0.1914 (2) &           0.183 (4) &          0.1885 (3) &          0.1823 (6) &          0.1828 (5) &          0.0078 (8) &          0.0486 (7) &          0.0007 (9) & \textbf{0.2083 (1)} \\
 $D_{6}$ &  0.6153 (8) &          0.6248 (5) &          0.6363 (3) &          0.6257 (4) &          0.6073 (9) &          0.6231 (6) & \textbf{0.6494 (1)} &         0.5216 (10) &          0.6488 (2) &          0.6183 (7) \\
 $D_{7}$ &  0.8093 (2) &          0.6387 (6) &          0.7163 (3) &          0.6076 (9) &          0.6362 (7) &          0.6339 (8) &          0.6946 (4) &         0.5408 (10) & \textbf{0.8263 (1)} &          0.6854 (5) \\
 $D_{8}$ & 0.8732 (10) &          0.8912 (7) &          0.8966 (4) &          0.8913 (6) &           0.897 (3) &          0.8923 (5) &          0.8775 (8) &          0.8984 (2) &          0.8747 (9) &  \textbf{0.909 (1)} \\
 $D_{9}$ & 0.2667 (10) &          0.5583 (5) &          0.5567 (6) &          0.5567 (6) &            0.56 (4) &          0.5467 (8) & \textbf{0.6247 (1)} &          0.5787 (3) &          0.5067 (9) &          0.5917 (2) \\
$D_{10}$ &   0.638 (8) & \textbf{0.8147 (1)} &          0.7667 (7) &          0.8013 (3) &          0.7974 (4) &          0.7904 (5) &          0.5454 (9) &         0.5188 (10) &          0.7827 (6) &          0.8071 (2) \\
$D_{11}$ &  0.4539 (4) &          0.4483 (6) &          0.4465 (7) & \textbf{0.5211 (1)} &          0.4484 (5) &          0.4809 (3) &          0.4207 (9) &          0.4309 (8) &         0.4119 (10) &           0.507 (2) \\
$D_{12}$ &     0.0 (9) &          0.3177 (2) &          0.3054 (3) &  \textbf{0.321 (1)} &          0.2988 (6) &          0.3023 (4) &             0.0 (9) &          0.0244 (8) &          0.1101 (7) &          0.3013 (5) \\
$D_{13}$ & 0.6163 (10) &          0.6628 (6) &          0.6634 (5) &  \textbf{0.675 (1)} &          0.6737 (2) &          0.6646 (4) &          0.6358 (8) &          0.6567 (7) &          0.6233 (9) &          0.6722 (3) \\
$D_{14}$ &  0.0216 (9) & \textbf{0.2744 (1)} &          0.2532 (5) &          0.2674 (3) &          0.2537 (4) &          0.2508 (6) &         0.0194 (10) &          0.0849 (8) &           0.086 (7) &          0.2723 (2) \\
$D_{15}$ &  0.9866 (5) &          0.9886 (3) &          0.9886 (4) &          0.9703 (9) &          0.9706 (8) & \textbf{0.9889 (1)} &          0.9783 (7) &         0.9591 (10) &          0.9813 (6) &          0.9887 (2) \\
$D_{16}$ &  0.2232 (9) &          0.6101 (4) &          0.6272 (2) &          0.6047 (6) &          0.5985 (7) &          0.6202 (3) &  \textbf{0.649 (1)} &         0.1938 (10) &          0.4262 (8) &          0.6061 (5) \\
$D_{17}$ &     0.0 (9) &          0.1786 (2) &          0.1568 (5) &          0.1625 (3) &          0.1578 (4) &          0.1416 (7) &          0.1489 (6) &          0.0756 (8) &             0.0 (9) & \textbf{0.1868 (1)} \\
$D_{18}$ &  0.1113 (9) &          0.5414 (6) & \textbf{0.5862 (1)} &          0.5751 (2) &          0.5659 (4) &          0.5749 (3) &            0.51 (7) &          0.059 (10) &          0.3838 (8) &          0.5428 (5) \\
$D_{19}$ &  0.376 (10) & \textbf{0.5188 (1)} &          0.4862 (4) &          0.5129 (3) &          0.4746 (6) &           0.484 (5) &          0.4124 (8) &          0.4345 (7) &          0.3998 (9) &          0.5133 (2) \\
$D_{20}$ & 0.0369 (10) &          0.2484 (3) &          0.2465 (4) & \textbf{0.2647 (1)} &          0.2406 (5) &          0.2206 (6) &          0.2005 (7) &          0.1598 (8) &          0.0708 (9) &          0.2509 (2) \\
$D_{21}$ &    0.0 (10) &          0.1209 (2) &          0.0712 (6) &          0.0562 (7) &          0.0516 (8) &          0.1157 (4) &           0.118 (3) & \textbf{0.1369 (1)} &            0.01 (9) &          0.0844 (5) \\
$D_{22}$ &    0.0 (10) &          0.1879 (3) &          0.1799 (4) & \textbf{0.2036 (1)} &          0.1786 (5) &          0.1879 (2) &          0.0659 (7) &           0.059 (8) &          0.0022 (9) &          0.1724 (6) \\
$D_{23}$ &     0.0 (9) &          0.1291 (4) &          0.0859 (7) &          0.2107 (2) &          0.0756 (8) &          0.0956 (6) & \textbf{0.2137 (1)} &          0.1081 (5) &             0.0 (9) &          0.1604 (3) \\
$D_{24}$ &    0.0 (10) &          0.2081 (5) &          0.1395 (6) &          0.2947 (2) &          0.1349 (7) &          0.0732 (9) &  \textbf{0.499 (1)} &          0.2443 (3) &           0.075 (8) &          0.2111 (4) \\
$D_{25}$ & 0.2816 (10) &          0.4883 (2) &          0.4771 (3) &          0.4708 (4) &          0.4664 (6) &          0.4444 (8) &  \textbf{0.489 (1)} &          0.4364 (9) &          0.4509 (7) &          0.4694 (5) \\
$D_{26}$ &    0.0 (10) &          0.2089 (3) &          0.1417 (6) &           0.181 (4) &          0.1344 (7) &          0.1439 (5) & \textbf{0.2216 (1)} &          0.1257 (8) &          0.0245 (9) &          0.2165 (2) \\
$D_{27}$ &    0.0 (10) &          0.1439 (6) &          0.1572 (3) & \textbf{0.1877 (1)} &           0.149 (4) &          0.1442 (5) &          0.1292 (7) &          0.0573 (9) &          0.0643 (8) &          0.1644 (2) \\
$D_{28}$ & 0.0033 (10) &          0.2853 (6) &          0.2986 (5) & \textbf{0.3486 (1)} &          0.2795 (8) &          0.3044 (4) &          0.3147 (3) &          0.2046 (9) &          0.3324 (2) &          0.2816 (7) \\
$D_{29}$ &  0.3038 (8) &          0.3608 (6) &          0.3846 (4) &          0.4543 (2) &          0.3036 (9) &          0.4303 (3) &          0.3788 (5) &         0.1782 (10) & \textbf{0.5984 (1)} &          0.3577 (7) \\
\midrule
Rank 1 (2) &       0 (2) &               4 (7) &               1 (2) &               7 (5) &      1 (1) &               1 (2) &               7 (0) &      2 (1) &               2 (3) &               \textbf{5 (9)} \\
Avg Rank &         8.5 &                 3.8 &   4.27 &  3.73&                 5.6 &                 4.9 &   5.33&                 7.3 &                 7.1 &   \textbf{3.53} \\
\bottomrule
\end{tabular}
}
\end{table}

\begin{table}[ht!]
\centering
\caption{MCC results (with ranks in parentheses) for each dataset and oversampling method across 29 benchmark datasets. Better performance is indicated by higher values and lower average ranks. The best-performing method for each dataset is shown in bold. ‘Rank 1(2)’ indicates the number of first- and second-place finishes, with the highest counts highlighted in bold.}
\label{tab:mcc_results}
\resizebox{\textwidth}{!}{
\begin{tabular}{c c c c c c c c c c c}
\toprule
 Dataset &          Original &                 ROS &               SMOTE &              bSMOTE &     ADASYN &              MWMOTE &               CTGAN &                GAMO &               MGVAE &                Ours \\
\midrule
 $D_{1}$ &           0.0 (7) &          0.0849 (4) &           0.096 (3) &          0.0569 (5) & 0.0973 (2) &          0.0474 (6) &         -0.0054 (9) &        -0.0059 (10) &         -0.0015 (8) & \textbf{0.1005 (1)} \\
 $D_{2}$ &       0.1646 (10) & \textbf{0.4259 (1)} &          0.3947 (4) &          0.3979 (3) & 0.3629 (7) &          0.3793 (5) &          0.2577 (9) &          0.3192 (8) &          0.3757 (6) &          0.3992 (2) \\
 $D_{3}$ &           0.0 (7) &          0.3923 (2) &           0.377 (3) &          0.3496 (5) & 0.3708 (4) &          0.3321 (6) &             0.0 (7) &         -0.064 (10) &         -0.0004 (9) & \textbf{0.4515 (1)} \\
 $D_{4}$ &        0.2435 (9) &          0.6974 (2) &          0.6238 (5) &          0.5846 (7) & 0.6238 (5) &          0.6781 (3) &          0.6539 (4) &          0.3674 (8) &         0.2429 (10) & \textbf{0.7401 (1)} \\
 $D_{5}$ &       0.0013 (10) &          0.1384 (2) &          0.1231 (4) &           0.128 (3) & 0.1226 (5) &          0.1207 (6) &          0.0123 (8) &          0.0356 (7) &          0.0022 (9) & \textbf{0.1515 (1)} \\
 $D_{6}$ &        0.5904 (8) &          0.6073 (4) &          0.6156 (3) &          0.6047 (5) & 0.5873 (9) &          0.6038 (6) & \textbf{0.6324 (1)} &         0.4824 (10) &           0.627 (2) &          0.6031 (7) \\
 $D_{7}$ &        0.8216 (2) &          0.6396 (6) &          0.7223 (3) &          0.6115 (9) & 0.6374 (7) &          0.6342 (8) &          0.6977 (4) &         0.5458 (10) & \textbf{0.8371 (1)} &          0.6877 (5) \\
 $D_{8}$ &        0.842 (10) &          0.8632 (6) &          0.8689 (4) &          0.8625 (7) & 0.8692 (3) &          0.8635 (5) &          0.8479 (8) &          0.8701 (2) &          0.8437 (9) & \textbf{0.8836 (1)} \\
 $D_{9}$ &       0.2638 (10) &          0.5574 (5) &          0.5545 (6) &          0.5545 (6) & 0.5583 (4) &           0.544 (8) & \textbf{0.6304 (1)} &          0.5933 (2) &          0.5056 (9) &          0.5911 (3) \\
$D_{10}$ &        0.6391 (8) & \textbf{0.8209 (1)} &          0.7736 (7) &          0.8078 (3) & 0.8047 (4) &          0.7969 (5) &         0.5398 (10) &          0.5473 (9) &          0.7864 (6) &          0.8139 (2) \\
$D_{11}$ &         0.496 (3) &          0.4406 (7) &           0.433 (9) & \textbf{0.5087 (1)} & 0.4371 (8) &          0.4675 (4) &          0.4533 (5) &         0.4158 (10) &          0.4486 (6) &          0.4975 (2) \\
$D_{12}$ &           0.0 (8) &          0.3309 (2) &           0.316 (4) &          0.3188 (3) & 0.3085 (5) &          0.3062 (6) &         -0.0002 (9) &        -0.0108 (10) &          0.1351 (7) & \textbf{0.3414 (1)} \\
$D_{13}$ &        0.4572 (8) &          0.4649 (4) &          0.4642 (5) &          0.4744 (3) &  0.475 (2) &          0.4639 (6) &         0.4489 (10) &          0.4518 (9) &          0.4594 (7) & \textbf{0.4776 (1)} \\
$D_{14}$ &        0.0461 (8) &          0.2358 (2) &          0.2126 (4) & \textbf{0.2433 (1)} & 0.2124 (5) &          0.2051 (6) &          0.0415 (9) &         0.0367 (10) &          0.1444 (7) &          0.2338 (3) \\
$D_{15}$ &        0.9845 (5) &          0.9869 (3) &          0.9869 (3) &          0.9657 (9) & 0.9661 (8) & \textbf{0.9872 (1)} &          0.9749 (7) &         0.9527 (10) &          0.9786 (6) &           0.987 (2) \\
$D_{16}$ &        0.2784 (9) &          0.5988 (4) &          0.6128 (2) &          0.5861 (7) & 0.5877 (6) &          0.5999 (3) & \textbf{0.6186 (1)} &         0.2035 (10) &           0.454 (8) &          0.5924 (5) \\
$D_{17}$ &      -0.0067 (10) &          0.1456 (2) &          0.1111 (6) &          0.1174 (4) & 0.1135 (5) &          0.0866 (7) &          0.1262 (3) &          0.0533 (8) &         -0.0056 (9) &  \textbf{0.148 (1)} \\
$D_{18}$ &        0.1851 (9) &          0.5473 (5) & \textbf{0.5856 (1)} &          0.5691 (3) & 0.5701 (2) &          0.5684 (4) &          0.5089 (7) &         0.0602 (10) &          0.4333 (8) &           0.537 (6) \\
$D_{19}$ &       0.4159 (10) &          0.4968 (2) &          0.4539 (4) &          0.4791 (3) & 0.4412 (6) &          0.4499 (5) &          0.4371 (7) &          0.4336 (8) &          0.4298 (9) & \textbf{0.5009 (1)} \\
$D_{20}$ &       0.0748 (10) &          0.2575 (2) &          0.2532 (3) & \textbf{0.2592 (1)} & 0.2477 (5) &           0.219 (6) &          0.1935 (7) &          0.1251 (8) &          0.1131 (9) &          0.2483 (4) \\
$D_{21}$ &          0.0 (10) &          0.1195 (2) &          0.0718 (6) &          0.0521 (7) & 0.0483 (8) &          0.1136 (3) &          0.1125 (4) & \textbf{0.1641 (1)} &          0.0062 (9) &          0.0836 (5) \\
$D_{22}$ &           0.0 (9) &          0.1829 (2) &          0.1748 (3) & \textbf{0.1842 (1)} & 0.1746 (4) &          0.1739 (5) &          0.0544 (7) &          0.0198 (8) &         -0.007 (10) &          0.1719 (6) \\
$D_{23}$ &           0.0 (9) &          0.1242 (4) &          0.0765 (7) &          0.2048 (2) & 0.0645 (8) &          0.0843 (6) & \textbf{0.2138 (1)} &          0.1105 (5) &        -0.0019 (10) &          0.1612 (3) \\
$D_{24}$ &          0.0 (10) &           0.202 (5) &          0.1258 (6) &          0.2925 (3) & 0.1215 (7) &          0.0594 (9) & \textbf{0.5081 (1)} &          0.3053 (2) &          0.0752 (8) &          0.2123 (4) \\
$D_{25}$ &       0.3107 (10) & \textbf{0.4481 (1)} &          0.4304 (4) &          0.4239 (6) & 0.4225 (7) &          0.3907 (8) &          0.4413 (2) &          0.3794 (9) &          0.4339 (3) &          0.4251 (5) \\
$D_{26}$ &           0.0 (9) &          0.2083 (3) &           0.137 (6) &          0.1714 (4) & 0.1318 (7) &            0.14 (5) &          0.2108 (2) &          0.1071 (8) &        -0.0022 (10) & \textbf{0.2258 (1)} \\
$D_{27}$ &           0.0 (9) &          0.1106 (5) &          0.1291 (3) & \textbf{0.1606 (1)} & 0.1188 (4) &          0.1067 (6) &          0.0852 (7) &        -0.0076 (10) &          0.0374 (8) &          0.1407 (2) \\
$D_{28}$ &       0.0044 (10) &          0.3114 (6) &            0.32 (3) & \textbf{0.3533 (1)} &  0.303 (8) &          0.3165 (5) &          0.3086 (7) &          0.1811 (9) &          0.3401 (2) &          0.3166 (4) \\
$D_{29}$ &        0.3367 (9) &          0.4022 (7) &          0.4198 (5) &          0.4668 (2) &  0.349 (8) &          0.4564 (3) &          0.4244 (4) &         0.1927 (10) & \textbf{0.6048 (1)} &          0.4067 (6) \\
\midrule
Rank 1(2) &             0 (2) &              3 (10) &      1 (1) &               6 (2) &      0 (3) &               2 (0) &               5 (2) &               1 (3) &               2 (3) &              \textbf{10 (5)} \\
Avg. Rank & 8.27 &                 3.7 &                 4.3 &                 4.0 &        5.6 &   5.23 &   5.47 &                 7.8 &   6.93 &   \textbf{3.07} \\
\bottomrule
\end{tabular}
}
\end{table}

\begin{figure*}[ht!]
\centering
\includegraphics[width=0.95\textwidth]{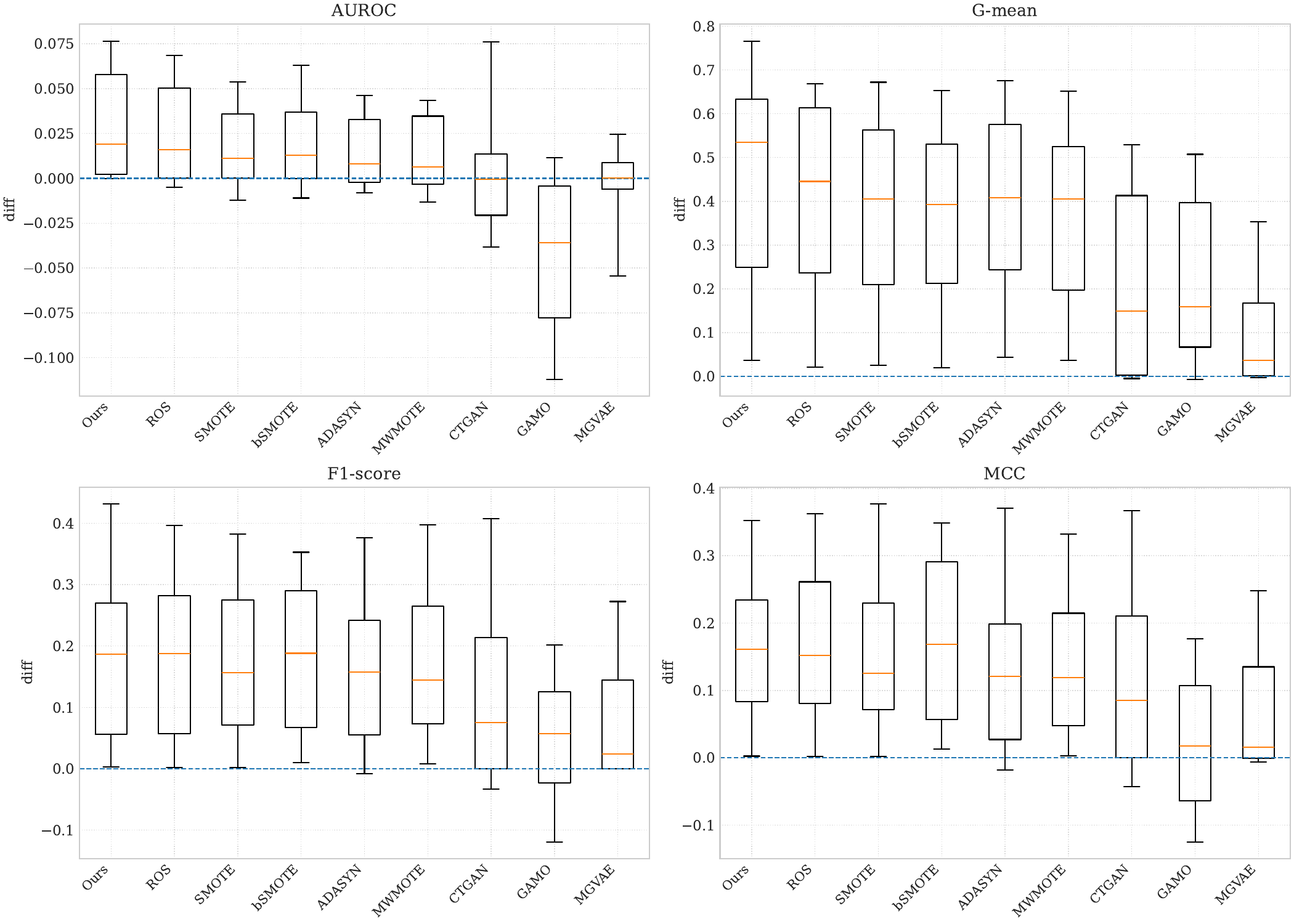}
\caption{Performance gains per metric relative to the Original baseline. Each boxplot illustrates the distribution of these performance differences across 29 real-world datasets.}
\label{fig:metric_diff_boxplot}
\end{figure*}

\vspace{0.5em}
\noindent \textbf{Gain over Original Baseline.} Figure \ref{fig:metric_diff_boxplot} presents boxplots of per-dataset performance differences between each method and the Original baseline for each evaluation metric. Our method achieves the largest and most stable improvements across all four metrics, with median gains outperforming other techniques. 

\vspace{0.5em}
\noindent \textbf{Statistical Significance.} 
To evaluate whether the observed performance gains of our proposed method are statistically significant against (1) the Original baseline and (2) the second-best performing method, ROS, we conducted paired t-tests using metric values from 29 datasets. As summarized in Table \ref{tab:t_test_results}, our method achieved statistically significant improvements across all four evaluation metrics against both benchmarks. Given its clear superiority over ROS, further comparisons with other methods are unnecessary. These findings substantiate the effectiveness of our approach in imbalanced classification.

\begin{table}[ht!]
\centering
\caption{P-values from paired t-tests comparing our proposed method with the Original baseline method and the ROS method (the second-best performer). Statistically significant results (\(p < 0.05\)) are highlighted in bold.}
\label{tab:t_test_results}
\begin{tabular}{ccc}
\toprule
Metric &  Original baseline & ROS method \\
\midrule
AUROC & $\mathbf{<0.001}$& $\mathbf{0.009}$ \\
G-mean & $\mathbf{<0.001}$& $\mathbf{0.010}$ \\
F1-score &$\mathbf{<0.001}$& $\mathbf{0.047}$  \\
MCC &$\mathbf{<0.001}$& $\mathbf{0.016}$ \\
\bottomrule
\end{tabular}
\end{table}

\subsection{Ablation Studies} \label{subsec: ablation}

We conduct a series of ablation studies to isolate the effects of key components in the proposed framework and to validate its robustness across varying experimental conditions. These studies include analyses of the regularization parameter $\lambda$, base classifier compatibility, comparative performance across different minority sample size, and empirical computational cost under varying input dimensionality and dataset size. 

\vspace{0.5em}
\noindent \textbf{Effect of the Regularization Parameter $\lambda$.} The regularization parameter $\lambda$ controls the trade-off between global distributional alignment via MMD and local boundary-aware regularization via triplet loss (see Equation~\ref{eq: total_obj}). We investigate the effect of varying $\lambda \in \{0, 0.001, 0.005, 0.01, 0.05, 0.1\}$ on overall classification performance, reporting average results across all 29 datasets in Table~\ref{tab:lambda_ablation}. The performance improves steadily as $\lambda$ increases from $0$ to $0.01$ across all four evaluation metrics, indicating that incorporating moderate boundary-aware regularization improves the informativeness of synthetic examples. However, for $\lambda > 0.01$, performance consistently declines, suggesting that excessively emphasizing local boundary structure undermines the global distributional alignment allocated by MMD. We therefore fix $\lambda = 0.01$ for all results in Section~\ref{subsec: main_results}. While this value is not optimal for every dataset individually, it provides a consistently strong performance across evaluation metrics. 

\begin{table}
  \centering
  \caption{Effect of regularization parameter $\lambda$ on average performance (SVM base classifier). Best results for each metric are bolded.}
  \begin{tabular}{c|ccccc}
    \toprule
    $\lambda$ & AUROC & G-mean & F1-score & MCC \\
    \midrule
    0     & 0.8878 & 0.7543 & 0.4998 & 0.4883 \\
    0.001 & 0.8880 & 0.7566 & 0.5009 & 0.4896 \\
    0.005 & 0.8887 & 0.7618 & 0.5014 &  0.4907 \\
    0.01  & \textbf{0.8891} & \textbf{0.7647} & \textbf{0.5036} & \textbf{0.4935} \\
    0.05  & 0.8884 & 0.7603 & 0.4937 & 0.4832 \\
    0.1   & 0.8822 & 0.7569 & 0.4792 & 0.4702 \\
    \bottomrule
  \end{tabular}
  \label{tab:lambda_ablation}
\end{table}

\begin{figure}[ht!]
    \centering
    \includegraphics[width=0.95\linewidth]{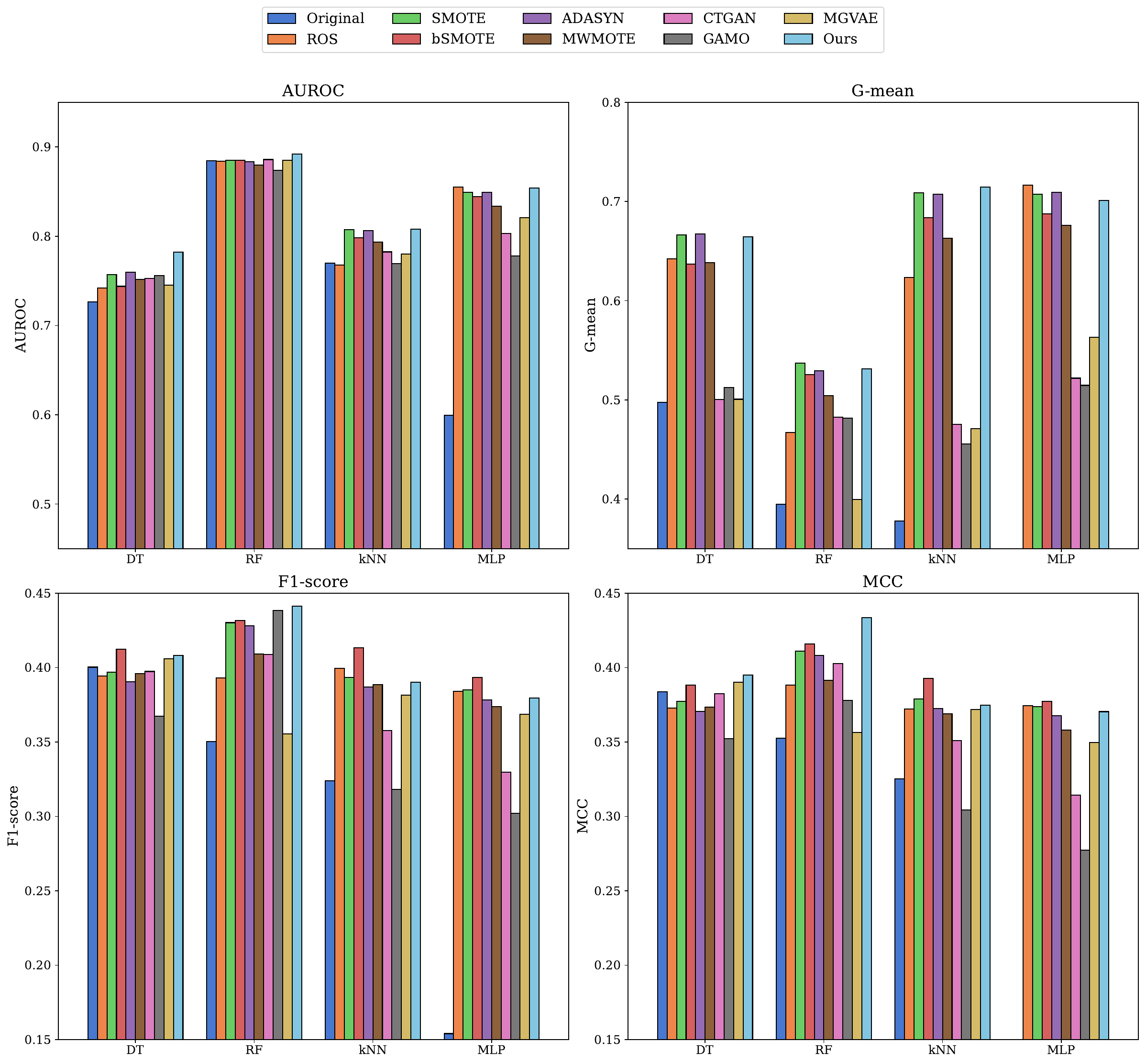}
    \caption{
    Average classification performance using four base classifiers—Decision Tree (DT), Random Forest (RF), $k$-Nearest Neighbors ($k$NN), and Multi-Layer Perceptron (MLP). Each bar shows the average metric achieved by a specific oversampling method over all datasets.
    }
    \label{fig:basemodels}
\end{figure}

\vspace{0.5em}
\noindent\textbf{Base Classifiers.} In  Section \ref{subsec: main_results}, we report experiments using a SVM classifier to validate our approach. To assess the general applicability of the proposed oversampling framework, we evaluate its performance in conjunction with a range of standard classifiers: Decision Tree (DT), Random Forest (RF), $k$-Nearest Neighbors ($k$NN), and Multi-Layer Perceptron (MLP). As depicted in Figure \ref{fig:basemodels}, the method consistently delivers the highest or near-highest mean values in AUROC, G-mean, F1-score, and MCC, averaged over 29 real-world datasets, regardless of which classifier is used. The consistent performance across models demonstrates that our method generates synthetic minority samples that enhance decision boundary learning across diverse classifiers. Moreover, the approach remains both effective and stable regardless of the classifier’s complexity or architecture. This confirms the practical versatility of our framework, making it suitable for diverse real-world applications where the choice of classifier may depend on task-specific requirements or computational constraints.

\begin{figure}[ht!]
    \centering
    \includegraphics[width=0.95\linewidth]{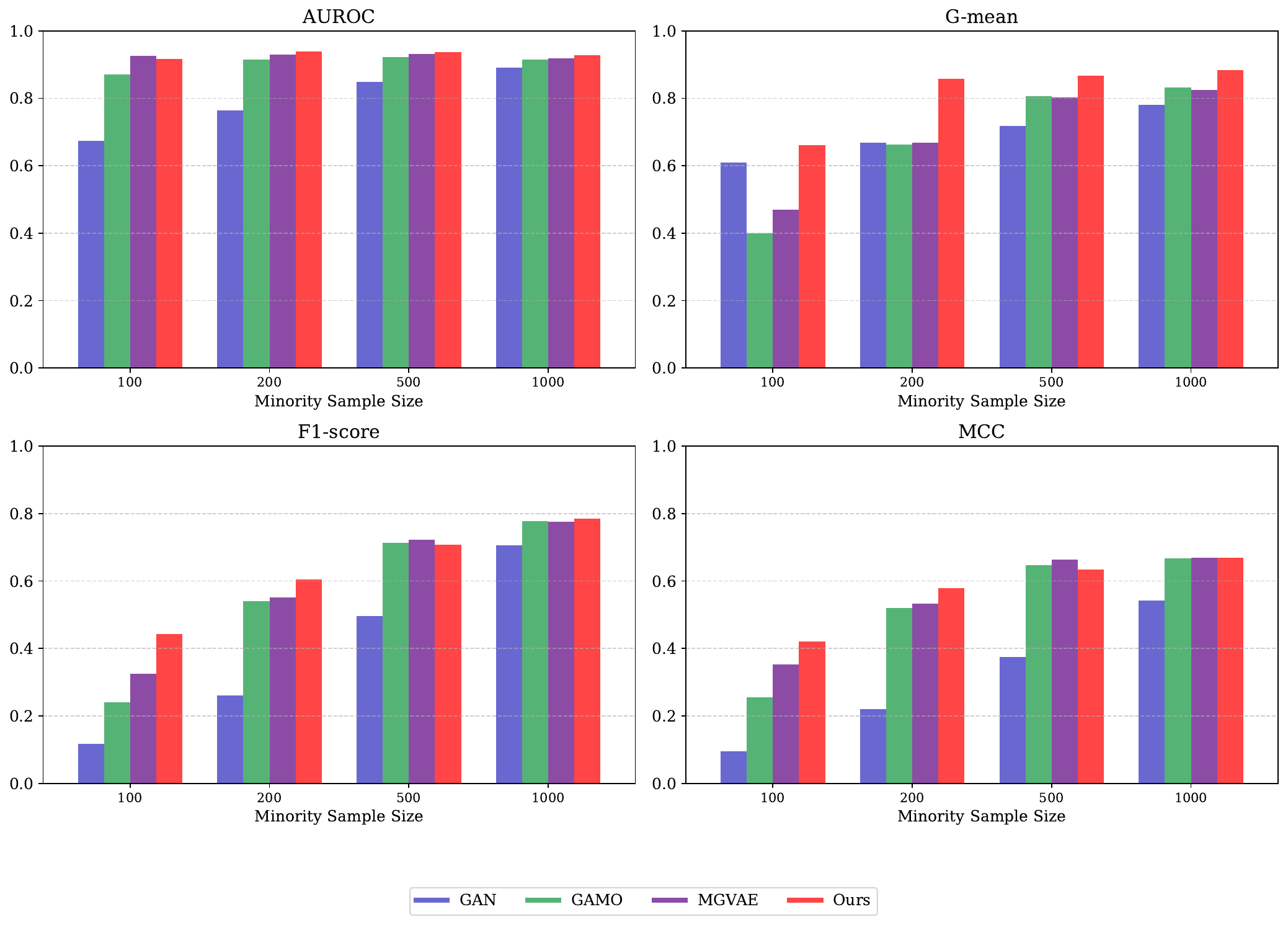}
    \caption{
    Average classification performance across varying $N_{\text{min}}$ with a fixed $N_{\text{maj}}$. Each bar represents the mean metric value over 10 independent runs for the generative methods. 
    }
    \label{fig:Nmins_ablation}
\end{figure}

\vspace{0.5em}
\noindent\textbf{Minority Sample Size.} To assess the robustness of our proposed oversampling framework under varying levels of minority class sparsity, we compare its performance against generative approaches on a simulated dataset. The dataset is generated from standard Gaussian distributions in a 50-dimensional space, with a fixed majority class size of $2000$ samples. Minority samples are similarly generated but with a mean shift of $0.3$, and their size varies across $N_{\text{min}} \in \{50, 100, 200, 500, 1000\}$. The simulations are repeated over 10 times to ensure reliability. Figure \ref{fig:Nmins_ablation} shows the average AUROC, G-mean, F1-score, and MCC across these $N_{\text{min}}$ values for GAN, GAMO, MGVAE, and our method. The results indicate that our approach consistently achieves highest or near-highest performance across all metrics, particularly in sparse regimes where GAN-based methods exhibit instability due to limited training data for realistic sample generation. As $N_{\text{min}}$ increases, the performance gap narrows, reflecting the improved capacity of generative models to capture minority distributions with more examples, yet our method maintains a competitive edge. This shows how our framework works better in very unbalanced situations by directly mapping many examples to the few, avoiding the need for many minority samples that generative models require. 

\begin{figure}[ht!]
    \centering
    \includegraphics[width=0.95\linewidth]{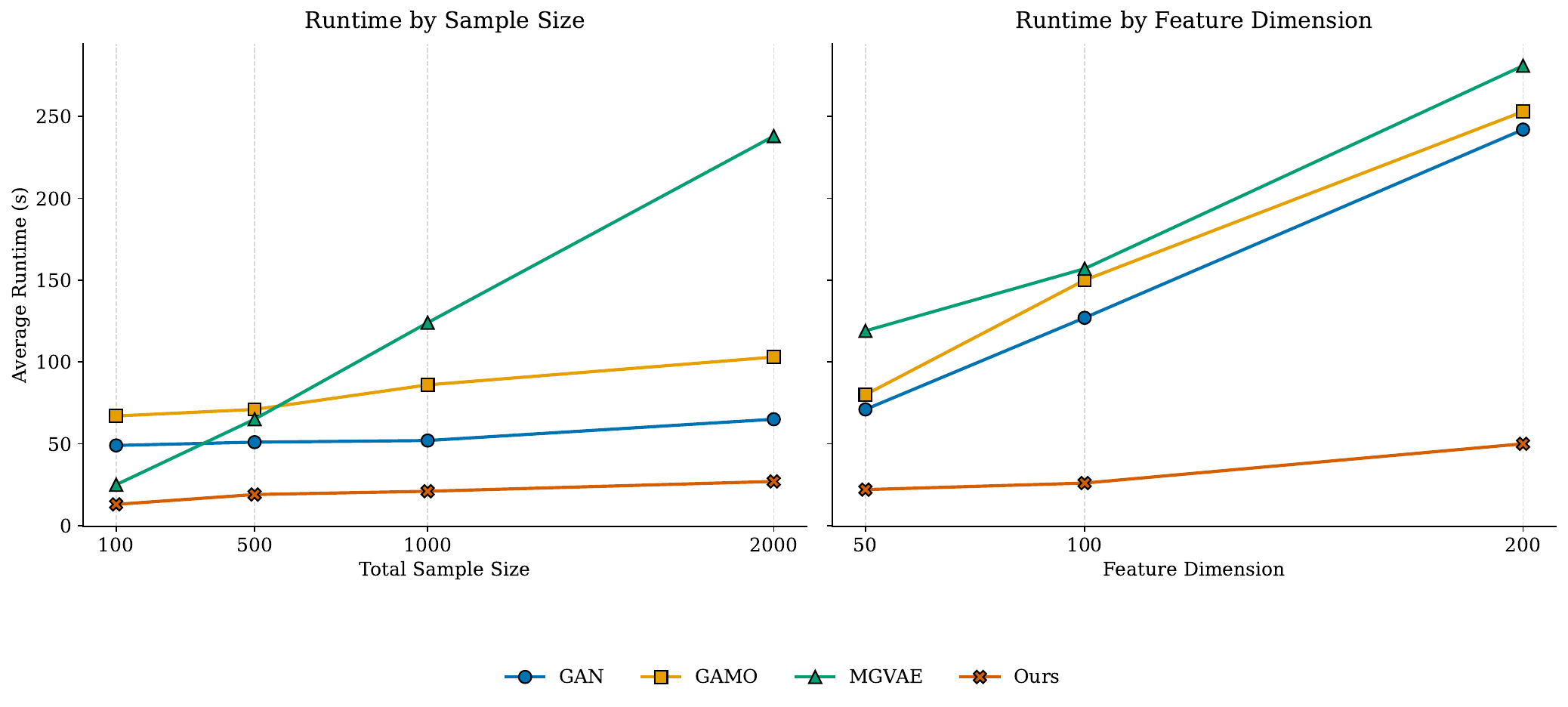}
    \caption{
    Average runtime (seconds) comparison of generative oversampling methods across total sample size and feature dimension on the Gaussian Blob simulation dataset described in Section \ref{sec:simulation_studies}. 
    }
    \label{fig:runtime_ablation}
\end{figure}

\vspace{0.5em}
\noindent \textbf{Computational Cost.} We evaluate the computational efficiency of the proposed method relative to generative baselines under varying sample sizes and feature dimensionalities. The experiments employ the same Gaussian Blob dataset described in Section \ref{sec:simulation_studies}. Figure \ref{fig:runtime_ablation} depicts the average runtime across methods as a function of $N_{\text{total}}$ and input feature dimension $d$. The results reveal that our method scales roughly linearly in both sample size and dimensionality, consistent with the complexity of computing the MMD and triplet loss. Compared to other generative oversampling models, the proposed approach demonstrates favorable computational efficiency. Non-adversarial optimization of our method circumvents the computational burden typical of GAN training. This advantage of the proposed framework underscores its practical utility in real-world datasets where dataset size and dimensionality can vary widely.  

\section{Conclusion} \label{sec: conclusion}
In this work, we have introduced a novel oversampling framework for imbalanced classification that learns a parametric transformation mapping majority samples to the minority distribution. By integrating a kernel-based Maximum Mean Discrepancy (MMD) with a triplet loss regularizer, our approach unifies global distributional alignment and local boundary-awareness in a non-adversarial manner. To validate its effectiveness, we conducted extensive empirical evaluations on 29 real-world benchmark datasets spanning diverse domains. We compared our method against classical interpolation methods (e.g., SMOTE, ADASYN) and state-of-the-art generative models (e.g., VAE- and GAN-based oversamplers) using four standard metrics: AUROC, G-mean, F1-score, and MCC. Across all metrics, our method achieved the best average rank, with consistent first- and second-place finishes on the majority of datasets. Paired t-tests confirmed that these improvements over both the Original baseline and the next-best methods are statistically significant.

We further demonstrate robustness across four base classifiers—Decision Trees, Random Forests, k-Nearest Neighbors, and Multi-Layer Perceptrons—showing best performance when swapping underlying classifiers. Ablation studies isolate the contributions of the MMD alignment and triplet-loss regularization, revealing that each component yields substantial gains in minority boundary learning. Finally, runtime analyses verify that our approach scales approximately linearly with sample size and feature dimension, outperforming generative methods in computational efficiency.

These comprehensive results underscore the versatility, stability, and practical utility of our proposed transport-map-based oversampling framework in addressing class imbalance across a wide range of settings.

Future research directions include extending the transformation map to multi-class and multi-label imbalance scenarios; investigating adaptive regularization strategies; and integrating the framework into end-to-end classifier pipelines to jointly optimize synthetic sample generation and predictive accuracy. Furthermore, theoretical analyses to establish convergence properties and distributional guarantees under complex data modalities represent promising avenues for deepening our understanding and broadening the framework’s impact.
\section*{Acknowledgments}
Suman Cha's work was supported by the National Research Foundation of Korea (NRF) grant funded by the Korean government (RS-2016-NR017145). 
Hyunjoong Kim’s work was supported by the IITP (Institute of Information \& Communications Technology Planning \& Evaluation)–ICAN (ICT Challenge and Advanced Network of HRD) grant funded by the Korea government (Ministry of Science and ICT) (IITP-2023-00259934) and by the National Research Foundation of Korea (NRF) grant funded by the Korean government (RS-2016-NR017145).

\bibliographystyle{plainnat}
\bibliography{reference}

\end{document}